\documentclass[letterpaper, 10 pt, conference]{ieeeconf}  

\IEEEoverridecommandlockouts                              
\usepackage[utf8]{inputenc}
\usepackage[T1]{fontenc}

\usepackage{graphics} 
\usepackage{epsfig} 
\usepackage{mathptmx} 
\usepackage{amsmath} 
\usepackage{amssymb}  

\usepackage[bookmarks=false]{hyperref}
\usepackage{cite}
\usepackage{algorithmic}
\usepackage{array}
\usepackage{stfloats}
\usepackage{url}

\title{
Addressing Cold Start in Recommender Systems with Hierarchical Graph Neural Networks
\author{Ivan Maksimov$^{1}$, Rodrigo Rivera-Castro$^{1}$ and Evgeny Burnaev$^{1}$
\thanks{Evgeny Burnaev was supported by RFBR grant 20-01-00203. Rodrigo Rivera-Castro was supported by the Mexican National Council for Science and Technology (CONACYT), 2018-000009-01EXTF-00154. 978-1-7281-6251-5/20/\$31.00 \copyright2020 IEEE }
\thanks{$^{1}$Skoltech, Moscow, Russia,
        {\tt\small ivan.maksimov@skoltech.ru}}%
}
}

\begin{document}
\maketitle
\thispagestyle{empty}
\pagestyle{empty}

\begin{abstract}
Recommender systems have become an essential instrument in a wide range of industries to personalize the user experience. 
A significant issue that has captured both researchers' and industry experts' attention is the cold start problem for new items.  
This work presents a graph neural network recommender system using item hierarchy graphs and a bespoke architecture to handle the cold start case for items. 
The experimental study on multiple datasets and millions of users and interactions indicates that our method achieves better forecasting quality than the state-of-the-art with a comparable computational time.

\end{abstract}
\section {Background} \label{sec:I_background}
A recommender system (recsys) is an efficient tool for matching items to customer interests. Generally, this system recommends items to users by leveraging already existing interactions between them. 
When the number of these interactions is big enough, it is called a "warm start" scenario, and the recsys performs well. 
However, often recsys faces a significant issue called "cold start" for items. 
This problem happens when a new item appears, and it only has a few interactions. 
Item cold-start recommendations are a challenge in video hosting, cinemas, social networks, e-commerce, food tech, fashion, and others. 
For example, YouTube is a video hosting facing the cold start problem for most of its recommendations as most of its content is new and becomes obsolete in a few days or weeks. 
These recommendations comprise up to 60 percent of the clicks on the home screen, \cite{youtube}. 
Netflix faces a similar challenge for almost 75 percent of all its video content, \cite{netflix}. 
Hence, achieving improvements for the cold start setting may yield a considerable business impact.

\section{Contributions} \label{contributions}

\begin{itemize}
    \item We introduce a novel data type for recommendation task, the multilevel item hierarchy graph;
    \item We present a new hierarchical graph embedding (HGE) algorithm able to address the cold start problem by exploiting the item hierarchy structure;
    \item For the HGE model, we improve a standard graph convolutional layer for a particular case of the item hierarchy graph. It leads to a decrease in the number of parameters from $O(\# items * log_2(\# items))$ to $O(\# items)$;
    \item Inside the HGE layer, we adapt the ReLU mechanism to make it possible to zero impact some item embeddings on the final graph embedding;
    \item We show that HGE yields superior performance in terms of precision@k and HR@k on medium and large-sized datasets with several item hierarchy levels;
    \item We show that HGE is robust even on small datasets with few item hierarchy levels;
    \item We show that HGE does not result in a considerable increase in computational time. We obtain only a 14\% increase in computational time compared with canonical matrix factorization under various conditions.
\end{itemize}

\section{Motivation} \label{sec:I_motivation}
Recsys are a fast-developing field. 
Numerous approaches appear each year. However, only a small portion of them yield significant impact, as shown in \cite{much_progress}. 
Broadly, there are three widely spread ways to overcome the cold start problem for items, \cite{cold_start, cold_item}. 
The first approach is to use a few interactions in collaborative filtering techniques. However, item embeddings become less stable if there are too few interactions, so the recommendations' quality suffers. 
The second approach for addressing the cold start consists of three stages. 
In the first stage, we fit any recsys with warm items containing a sufficient number of interactions.
In the second stage, we train a machine learning regression model to predict item embeddings from meta information. 
Finally, we predict their embeddings with the model's help from stage two with the meta-information for cold items. 
This approach's disadvantage is that such a model does not directly predict a target for cold items. 
It predicts either items and user embeddings or only one of the two of them. 
Our hope is then that the dot product of such predicted embedding will yield a consistent result. 
The third approach is to incorporate a mechanism for addressing cold start inside the collaborative filtering model as an additional linear term that forms item embedding. 
These methods have a lot in common as they lever meta-information from items. A widely spread use of such types is item hierarchy. 
Companies in e-commerce, retail, fashion, and many other use item catalogs and categorize items. 
However, most of the existing methods for addressing the cold start, including the above mentioned, treat it as categorical data with no hierarchy structure, \cite{cold_item}. 
We claim that developing an algorithm designed to work with a specific item hierarchy data structure effectively may significantly improve the cold start scenario's model performance.

\section{Innovation} \label{I_innovation}
This work shows that implementing a unique hierarchical graph-based recsys can improve the recommendations quality for top-N recommendations under the cold start for the items scenario without a considerable increase in computational time. 
The work contributes to the growing area of graph neural network research twofold. 
First, we introduce a novel data type for recsys, the multilevel hierarchical item graphs. 
Second, we present a bespoke graph convolutional neural network to process it.
In \autoref{fig:4.3:algorithm}, we present our approach for handling hierarchical item graph information. 
The algorithm efficiently handles the input graph hierarchical structure using a novel graph convolutional mechanism. 
We show that this mechanism considerably reduces the number of layers' parameters compared with classic graph neural networks. 
Further, the proposed algorithm can benefit from complex multilevel hierarchy structures and be robust enough to reduce hierarchies to one level or category.
We validate the approach by testing it on five datasets for item recommendations using metrics for hit rate@$k$ and precision@$k$. 
We consider several values for $k$, and all of them are consistent with our conclusions. 
The results show that the proposed algorithm performs better than the canonical matrix factorization method, hybrid models, and modern graph-based recsys when multilevel item hierarchy information is present and not worse than when hierarchical information consists of only one category.

\section{Literature Review} \label{LR}

\subsection{Heuristics} \label{sec:heuristics}
According to authors such as \cite{much_progress, various_mf, expl_rules, practical_1, baselines, au}, simple recsys without machine learning behind them can perform well in some cases. 
Random recommendations are one of the most straightforward yet powerful recommendation systems in cold start cases as they are not dependent on the number of available interactions. 
Thus, even if only one interaction with an item exists, a random recsys can recommend it. 

\subsection{Matrix Factorization Based Methods} \label{MF}

\subsubsection{Canonical Matrix Factorization}  \label{vMF}
Matrix factorization (MF) is a classic yet compelling technique. 
Many industrial applications, \cite{intro_mf, phdthesis} make use of it, due to its scalability and various extensions that help to incorporate side information into the matrix factorization, \cite{various_mf}. 
The classic MF model presented is defined as $R = W_1 * W_2^T$. 
It has a structure where $R$ is the user-item matrix, $W_1$ is the user embedding, and $W_2$ the item embedding.

This model assumes that the resulting score is a dot product between a user and an item embedding. An embedding is a low-dimensional representation of a user or an item. With stochastic gradient descent or other gradient descent methods, it is possible to learn the model. This model does not handle external features into account. 

The simplicity of MF has many advantages, \cite{mf_survey}. For example, it is incredibly scalable and fast due to matrix multiplication. 
The operation is faster than an iterative for-loop. 
Moreover, we can parallelize it. 
Secondly, there is a straightforward way of applying regularization to the model.
For this, we add an L2 regularization on $W_1$, $W_2$ matrices' norm.
Interestingly, we can write a matrix factorization as a neural network with two separate embedding layers. 
This property is fundamental. 
It enables the possibility to add external features, apply neural-based methods, and other classic improvements for neural networks such as batch normalization, cyclical learning rate, data augmentation, and others. 
Also, user and item embeddings derived from MF have an excellent property; similar items are close to each other in embedding space.

\subsubsection{Alternating Least Squares}  \label{als}
Alternating Least Squares (ALS) is a matrix factorization method widely adopted in the industry, \cite{als_experiment}. 
ALS has the same model structure as the canonical MF. 
Due to its simplicity, speed, and ability to train almost online, it enjoys significant popularity, \cite{als_ranking}. 
It also has relevant properties such as weights for instances and regularization, and we define it as

\begin{equation*}
\begin{split}
    L_{als} &= \sum_{u, i}{c_{ui}*(r_{ui} - x_u^T * y_i)^2} \\
&+ \lambda_x * \sum_u{||x_u||^2} + \lambda_y * \sum_i{||y_i||^2},
\end{split}
\end{equation*}\label{eq:als}

where $c_{ui}$ is the weight of element $ui$ in the user-item matrix, $r_{ui}$ is the binary flag, whether an element at position $ui$ is not zero, $x_u$ is the user embedding, $y_i$ is the item embedding, and $\lambda_x$ and $\lambda_y$ are the regularization terms.

Indeed, ALS is not a separate algorithm but a way to optimize the above function. 
ALS optimizes it in an alternating way. 
First, we fix user embeddings. 
With fixed user embeddings, $L_{als}$ is convex. 
Then we make a gradient step for item embeddings to update them. 
After that, we do the same operation with item embeddings and continue this alternating operating until convergence. 
\subsection{Hybrid Matrix Factorization}  \label{hMF}
There exist several ways to incorporate external features in MF, \cite{hybridmf, hybridsvd, nnmf}, and improve model performance in the cold start case. 
As with the canonical matrix factorization, we can describe it as a simple neural network, \cite{mf_survey}. 
One extension is the hybrid matrix factorization (HybridMF) to address the cold start by incorporating item features in the model. 
HybridMF uses a simple user and item embedding neural network with its dot product \cite{hybridmf}. 
We can define it as

\begin{equation*}
r_{ui} = (x_u + f_i \Theta_u)^T \cdot (y_i + f_i\Theta_i) + b_u + b_i.
\end{equation*} 

The following architecture parts differentiate HybridMF from the canonical matrix factorization. 
Firstly, we add a separate additional input for item features $f_u$. 
We combine item embedding with feature embedding by a sum operation, \cite{nnmf2, nnmf3}. 
The same logic holds for user features $f_u$. 
Also, user and item biases $b_u$, $b_i$ are added. 
More advanced ways of incorporating external features may yield minor improvements \cite{inbook}. 
However, researchers are undecided if more advanced techniques for adding external features such as neural-based models bring improvements, \cite{much_progress}. 
Thus we decided to include a stable and straightforward way of incorporating side information inside the MF model.

\subsection{Light Factorization Machines} \label{LightFM}
Light Factorization Machines (LightFM) is a powerful model that incorporates collaborative and content information, \cite{fm}. 
An essential property of LightFM is its ability to perform at least as well as simple content-based methods in the cold start and low-density cases, \cite{lightfm_lyst}. 
Indeed, as LightFM incorporates content information, it is well-suited for cold case scenarios. 
\cite{lightfm_lyst} shows that when little information about collaborative filtering and metadata is available, LightFM can outperform both content-based methods and collaborative filtering. 
We can construct the LightFM model as

\begin{equation*}
r_{ui} = \sigma(q_{u}*p_{i} + b_u + b_i),
\end{equation*} 

where $r_{ui}$ is the predicted probability, $\sigma$ is the sigmoid function, $q_{u}$ and $p_i$ are the user and item embeddings. 
Similarly, $b_u$ and $b_i$ are the biases for users and items.

We construct the user and item embeddings, $q_u$, and $p_i$, by summing up both collaborative filtering and content features' latent vectors. 
The same logic holds for the bias terms, $b_u$, and $b_i$. This property means that when only ids for users and items are present, LightFM reduces to the standard matrix factorization model. 
In the cold start scenario, we can represent items and users' embeddings with the user and item features. 
Thus, LightFM is capable of addressing the cold start problem.

\subsection{Graph-based Recommender Systems} \label{graph}
There are multiple Graph Neural Networks (GNN) proposals in the literature. 
Given the myriad of variants for GNN, we can find a good overview in \cite{gnn_survey, gnn_comparison}. 
Usually, these variants differ by graph types, training methods, and propagation step, \cite{gnn_overview}.
In the case of recsys, we consider mostly undirected graphs with node label information. 
Indeed, the user-item graph follows this structure as well as the proposed item hierarchy graph. 


The first propagation scheme to gain significant popularity among practitioners was probably the spatial convolutional aggregator, \cite{hamilton2017inductive, dyn_gcn}. 
It used an assumption that node embedding is an embedding of its neighbors and itself. 
We define the aggregation and update steps as

\begin{align*}
    h_{Nv}^t &= \text{AGGREGATE}(h_{u}^{t-1}, \forall u \in N_v) \\
h_{v}^t &= \sigma(W_t * [h_v^{t-1}||h_{Nv}^t ]),
\end{align*}

where $h_{Nv}^t$ is an embedding of neighbours of node $v$, $h_{v}^t$ is the embedding of node $v$, $[h_v^{t-1}||h_{Nv}^t ]$ is a stacked matrix of node and its neighbors embeddings, and AGGREGATE is an aggregation function. 
In its simplest form is a mean or sum.
One improvement is performing the convolutions in a spectral field or adding an attention mechanism, \cite{dual_graph_attention}. 
The convolution operation yields better results than the gated update mechanism, \cite{kgcn, kgat, conv_gnn, mc_gnn}. 

Graph convolutional networks are incredibly competitive in the cold start case. 
\cite{gcmc} treats the recommendation task as a link prediction in the bipartite graph. 
If we consider the canonical MF as a linear encoder-decoder model, the only difference is the encoder-decoder mechanism. 
Indeed, \cite{gcmc} benefits from additional knowledge about the local node neighborhood.

Similarly,  it shares weights between the embeddings for users and items through local convolutions for all nodes. 
This mechanism helps to improve speed but makes it more prone to overfitting. 
We can describe the weight sharing mechanism as

\begin{align*}
h_{i}^t &= \text{AGGREGATE}( \frac{1}{c_{ij}} W_r * h_j^{t-1}, \forall j \in N_i) \\
u_{i}^t &= \sigma(W * h_{i}^t),
\end{align*}

where $h_{i}^t$ is the embedding of node $i$, $c_{ij}$ is the normalization constant equal to the number of neighbours $|N_i|$, $W_r$ is the learnable matrix, and $u_i^t$  is the user embedding.
  
The community has recently looked into hierarchical graph pooling, \cite{implicit_hierarchical_structures, gnn_implementation, gnn_scalability}. 
These studies still use a user-item graph but propose a unique mechanism for node pooling. 
For example, \cite{implicit_hierarchical_structures} learns a differentiable soft cluster assignment for nodes. 
Our work differentiates itself by using a different type of graph. 
It uses a hierarchical graph, instead of a simple graph as in \cite{implicit_hierarchical_structures} or other works and a bespoke graph convolutional operator. 
Moreover, \cite{implicit_hierarchical_structures} is designed for \textit{entire} graph classification and not for recommendation.

\section{Hierarchical Graph Embedding}
\begin{figure}[ht]
    \centering
    \includegraphics[width=\columnwidth]{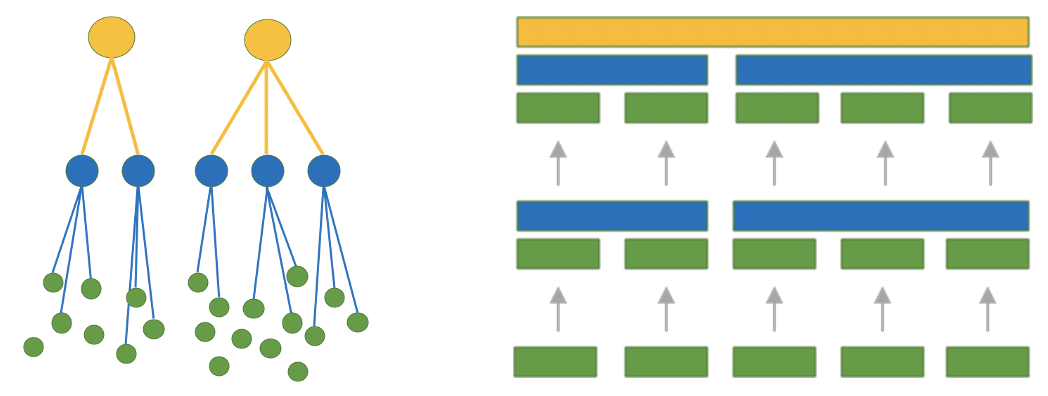}
    \caption{Data Structure and Down-Top Convolutions}
    \label{fig:4.1:data}
\end{figure}

We propose to use item neighbors' information in each hierarchy level to obtain more stable embeddings as depicted in \autoref{fig:4.1:data}. 
Our method models down-top convolutions to obtain a feature map of each hierarchy level.

We define the user-item matrix $M$. 
From it, we can get user and item embeddings in the canonical matrix factorization form. 
Let $I$ be the item embeddings matrix and $U$ the user embedding matrix. 
The input for the HGE layer is computed in the previous neural network layer item embedding $I$. 
We introduce $G$ as our convolution over the graph. 
As the item hierarchy graph at each level is a block diagonal matrix, we can simplify the graph convolution mechanism described in \cite{kgcn, gnn_scalability}. 
Hence, we define our convolution as 

\begin{equation*}
    G*W = G*W_1*W_2^T.
\end{equation*}

It consists only of a matrix multiplication mechanism and lets us factorize the matrix $W$ into two matrices $W_1$ and $W_2^T$. 

The second part of the HGE architecture is the activation function. 
We use a rectified linear unit (ReLU) activation function, $\textrm{ReLU} = \textrm{max}(0, x)$.
ReLU, by construction, has a helpful ability to zero negative inputs. 
It makes the final embedding more sparse and less prone to overfitting, \cite{relu}. 
Similarly, we use SoftMax and apply it to the matrix in a row-wise manner.

As a data structure consisting of a multilevel item hierarchies graph is present, we need to let features constructed at lower levels flow to the end of the neural network. 
An efficient way for it is to apply skip-connections, \cite{resnet}. 
The skip connection for any abstract neural layer $\textrm{layer}(x)$ is just adding input features $x$ to the output of the layer, $x + \textrm{layer}(x)$.

We can compute the HGE as
    
\begin{align*}
\textrm{HGE(I, G)} &= I*\textrm{SoftMax}(ReLU(G*W_1*W_2^T) \\
I &= I + HGE(I, G).
\end{align*}

$I$ is the item embedding matrix, $G$ is the graph adjacency matrix, and $W_1, W_2^T$ are learnable matrices.

\begin{figure}[ht]
    \centering
    \includegraphics[width=0.6\columnwidth]{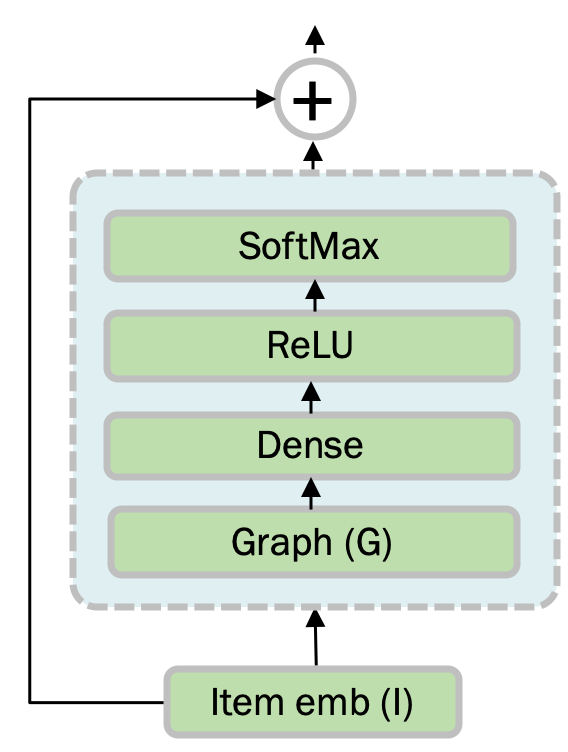}
    \caption{The Hierarchical Graph Embedding Layer}
    \label{fig:4.2:hge layer}
\end{figure}

The HGE layer depicted in \autoref{fig:4.2:hge layer} performs as follows. First, it gets item embeddings $I$ and an item hierarchy graph from a particular item hierarchy level $G$ as an input. 
Then, it applies a graph convolution. 
According to this operation, the resulting item embeddings are a weighted sum of input item embeddings. 
After that, it applies the ReLU activation function to bring non-linearity to the model and zero some values in the item embedding matrix. 
Next, it applies SoftMax in row-wise order to convert the item embedding into weights. 
It multiplies an initial input item embedding $I$ by the output of all the described above steps. 
Finally, it makes use of the skip-connection mechanism.

\begin{figure}[ht]
    \centering
    \includegraphics[width=\columnwidth]{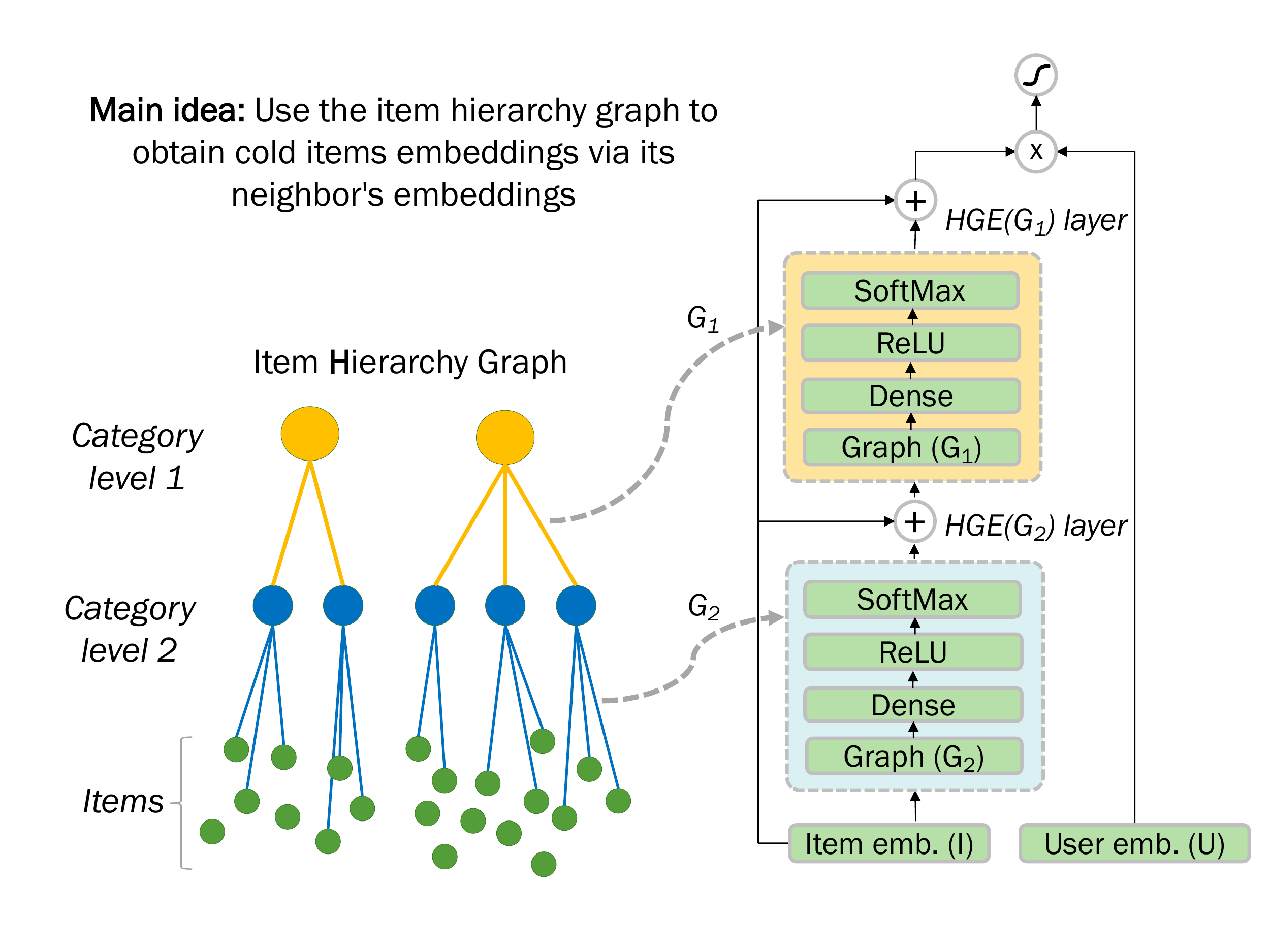}
    \caption{Overview of the Hierarchical Graph Embedding Model Structure}
    \label{fig:4.3:algorithm}
\end{figure}

Let $K$ be the number of categories at one item hierarchy level, $h$ be a hidden dimension size. 
The HGE layer has only $(I+K)*h$ parameters. 
In a worst-case scenario, when the item category graph is a binary tree $K$, it is equal to $log_2(I)$, and $h$ is a constant value. 
Asymptotically, the HGE layer has  $O((I+K)*h) = O(I + log_2(I)) = O(I)$ learnable parameters. 
Thus, asymptotically it does not make the entire proposed model more complex. 
Indeed, to obtain item embeddings via the canonical matrix factorization, we need $O(I)$ parameters. 
To apply the HGE layer to them, we use $O(I)$ parameters additionally. 
Thus $O(I) + O(I)$ = $O(I)$. 
So, asymptotically, MF with the HGE layer has the same parameters as its canonical form.  
We present an overview of the full model in \autoref{fig:4.3:algorithm}. 

In order to do this, we transformed an elaborate item hierarchy graph into several graphs. One item hierarchy layer corresponds to one item-item graph. 
It is much easier to apply the HGE layers to model a complex multilevel structure. 
Indeed, to do this, each HGE layer is dependent on its own separate item hierarchy graph. 
We use the skip-connection mechanism to let the information flow from the first layers to the last layer, as described in \cite{resnet}. 
We define the whole model as

\begin{equation*}
I = I + HGE_2(I + HGE_1(I, G_1), G_2).
\end{equation*}

Here, $G_1$ is the graph adjacency matrix at the first hierarchy level, and $G_2$ is a graph adjacency matrix at the second hierarchy level.

As HGE is an extension of the canonical matrix factorization, we will compare it in computational speed and against the canonical MF. 
For this, we set the following hypotheses. 
First, HGE outperforms all baselines in HR@k and PR@k when rich item hierarchy information is present.
Second, HGE performs at least not worse than baselines, even on small datasets with few hierarchy levels. 
Third, HGE yields competitive training time in comparison to baselines. 
Fourth, HGE item embeddings hierarchically produce clusters. 
The lower level category clusters are inside higher-level category clusters. 
Fifth, the more items per category in the data, the better HGE performs in HR@k, PR@k.

\section {Datasets Description} \label{datasets_description}
Our comparison uses datasets from different industries covering movies, songs, retail. 
The list of datasets includes Movielens 100K, \cite{movielens}, Movielens 1M, \cite{movielens}, Amazon Reviews, \cite{amazon}, Retail Rocket, \cite{dumn}, X5 Retail Hero, \cite{x5} and the Hotel Booking demand dataset, \cite{hotel}. 
We did not consider many widely used datasets because they do not provide item category information. 
As rating prediction is out of the work scope, we only consider the setting with binary feedback. 

We binarized the Movielens and Amazon datasets with a threshold rating value of 3, i.e., we set higher ratings to 1 and lower ratings to 0. 
We use only categorical, hierarchical, item features as item meta information. According to the methodology described in \cite{hybridsvd}, we ensured that each user and each item has at least five interactions. 
We present an overview of the datasets' main characteristics in \autoref{tab:data_sets}.

Further, we split the datasets into two groups. 
The first group consists of small datasets and not rich in item hierarchy information. 
We aim to show that the proposed method is robust with small data and little information for these datasets. 
The second group of datasets represents big datasets with rich multilevel item hierarchy. 
For this group, we show that the proposed model significantly outperforms all baselines.

\begin{table}[ht]
\centering
\caption{\textsc{datasets description}\label{tab:data_sets}}
\resizebox{0.48\textwidth}{!}{%
\begin{tabular}{lcccc}

Dataset & \# Users    & \# Items   & \# Interactions & \# Hierarchy Levels \\ \hline
MovieLens 100K  &  3K   & 2K     & 100K & 1  \\ \hline
MovieLens 1M &   6K   &    4K     & 1M & 1    \\ \hline
X5 Retail Hero &  122K    &    43K     &   2.5M  & 4   \\ \hline
Amazon &   530K   &  1.2M    &  10M & 3  \\ \hline
Retail Rocket &   1.4M   &    420K     &  12.2M & 3 \\ \hline
\end{tabular}
}
\end{table}

\section {Evaluation methodology} \label{evaluation_metrics}

\begin{figure}[ht]
    \centering
    \includegraphics[width=\columnwidth]{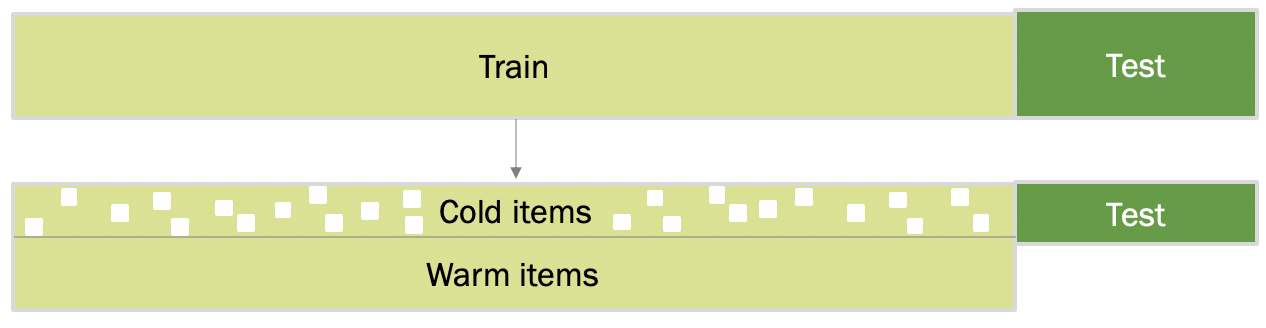}
    \caption{Proposed Validation Scheme for Experiments}
    \label{fig:3.1:validation_scheme}
\end{figure}

The goal of the model is to perform well in cold start cases. 
For this, the test data should consist of cold start items. 
In order to construct such test dataset, we follow the splitting logic reported in \cite{hybridsvd} with minor modifications shown in \autoref{fig:3.1:validation_scheme}. Finally, we construct the dataset in the following manner. 
First, we reserve the last two weeks for testing.
Second, we perform an 80\%/20\% partitioning of the list of all unique items.
Third, we mark items from a 20\% partition as cold items.
Fourth, for each item from a 20\% partition, we downsample user-item interactions on the train set up to 1\% of interactions to imitate the cold start problem.
Fifth, we measure the test metrics only for a 20\% partition.

By construction, such a train test split ensures to consider the item cold-start problem, \cite{hybridsvd}. 
Thus, we can precisely evaluate the performance of the model under these circumstances. 
We conduct experiments for top-N recommendations under a cold start scenario. 
Each experiment starts with hyperparameter tuning with $n=10$ fixed. 
After optimal hyperparameters are selected, we provide a more comprehensive range for n for final evaluation. 
For all methods, we evaluate an embedding size from 20 to 200 with a step of 20. 
Similarly, we explore a learning rate from 1e-5 to 1e-1 with a scale of x10.

To measure model performance, we consider the top-N recommendation task, \cite{metrics}. 
In particular, we use HR@k, \cite{much_progress}, and PR@k, \cite{handbook}. 

HR@k stands for the hit rate among the first $k$ recommendations. 
It shows what percent of users bought at least one recommended item among the top-k. 
PR@k stands for precision among the first $k$ recommendations. 
This metric assesses which percentage of items among the top-k recommended was relevant. 
We define HR@k as

\begin{equation*}
\textrm{HR@k} = \frac{1}{|U|} \sum_i {\textrm{[tk]}_i}.
\end{equation*}\label{eq:hrk}

For this, $tk$ represents that at least one of the top-k recommended items was relevant for user $i$. 
Similarly, we outline PR@k as 

\begin{equation*}
\textrm{PR@k} = \frac{1}{|U|} \sum_i { \frac{\textrm{ntkr}_i}{\textrm{ntk}_i}}.
\end{equation*}\label{eq:prk}

Here, $ntkr$ is the number of top-k recommended items relevant for user $i$, and $ntk$ is the number of top-k recommended items for user i.
  
With minimal values, $n=1$, these evaluation metrics are volatile, but with substantial values such as $n>200$, the model neglects its business value, \cite{metrics}. 
The common value of $n$ in industrial applications varies from 5 to 20, \cite{metrics}. 
To get a stable result, we provide results for $n$ equal to 10 and 20.

\section {Baselines} \label{sec:baselines}
We define as baselines the following methods, Random recommendation (Random), Canonical Matrix Factorization (MF), Alternating Least Squares (ALS), Hybrid Matrix Factorization (HybridMF), LightFM, Graph Convolution Matrix Completion (GC-MC).

We choose Random recommendations, MF, and ALS as baselines from the classic recsys literature. 
HybridMF helps us compare the proposed model with a classic model designed for the cold start case, \cite{hybridmf}. 
We consider LightFM as it is widely used in the industry and shows superior performance for real-world datasets under cold start conditions, \cite{lightfm_lyst}. 
We also compare HGE against GC-MC with features,  \cite{gcmc}. 
It is a graph-based model considered state of the art in a cold start setup for several classic datasets.

\section {Results} \label{Results}
\begin{figure*}
\begin{center}
\centerline{\includegraphics[width=\textwidth]{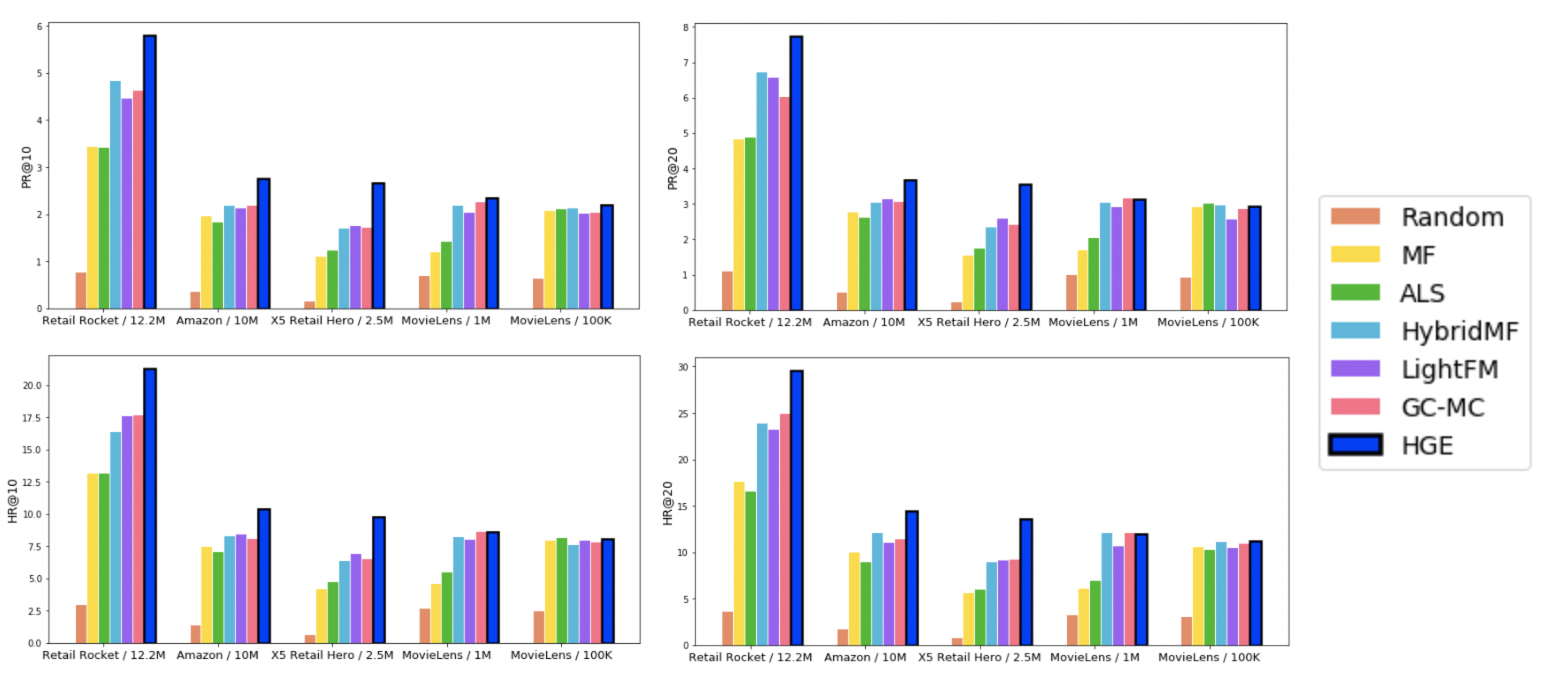}}
\caption{Precision@k (top) and HR@k (bottom) comparison between models and datasets for $k$ equal to 10 (left) and 20 (right). 
HGE outperforms baselines for medium and large datasets with a multilevel item hierarchy graph. 
Datasets from left to right: Retail Rocket, Amazon, X5 Retail Hero, Movielens 1M, Movielens 100k}
    \label{fig:5.1:ablation_study}
\end{center}
\end{figure*}

We present the results of the experiments in \autoref{fig:5.1:ablation_study}. 
We sort the datasets in the figure based on the number of user-item interactions, from the largest to the smallest.

As we can see from \autoref{fig:5.1:ablation_study}, HGE is not worse than baselines for all datasets. 
In particular, HGE is significantly better for three out of five datasets, Retail Rocket, Amazon, and X5 Retail Hero, on all metrics. 
For MovieLens 1M, it shows comparable performance with HybridMF and GC-MC. 
MovieLens 100K, four methods, MF, ALS, HybridMF, and HGE, have a comparable performance for the smallest dataset. 
LightFM is significantly worse than these four methods, probably due to overfitting on a small dataset.

HGE outperforms baselines significantly on large datasets with a considerable number of item hierarchy levels. 
The first three datasets, where HGE has shown a superior performance, are large and have multiple item hierarchy layers. 
For example, Retail Rocket has 12.2M interactions and three levels,  Amazon has 10M  interactions and three levels, and X5 Retail Hero has 2.5M interactions and four levels. 
Also, HGE is robust even for small datasets with few or even one level of item hierarchy on all baselines with MovieLens 1M and MovieLens 100K.

HGE produces clustered item embeddings according to item categories. 
Indeed, a useful property of embedding that we hope to obtain is that similar items are close in the item embeddings space. 
As items from one category are similar, we hope to have categorical clusters in the item embedding space. Often, conventional methods in the literature do not have this property \textit{directly by construction} or by an optimization technique. 
In HGE, when we pass item embeddings through the HGE layer, the item embeddings from one group are pushed closer to each other.

\begin{figure*}
\begin{center}
\centerline{\includegraphics[width=\textwidth]{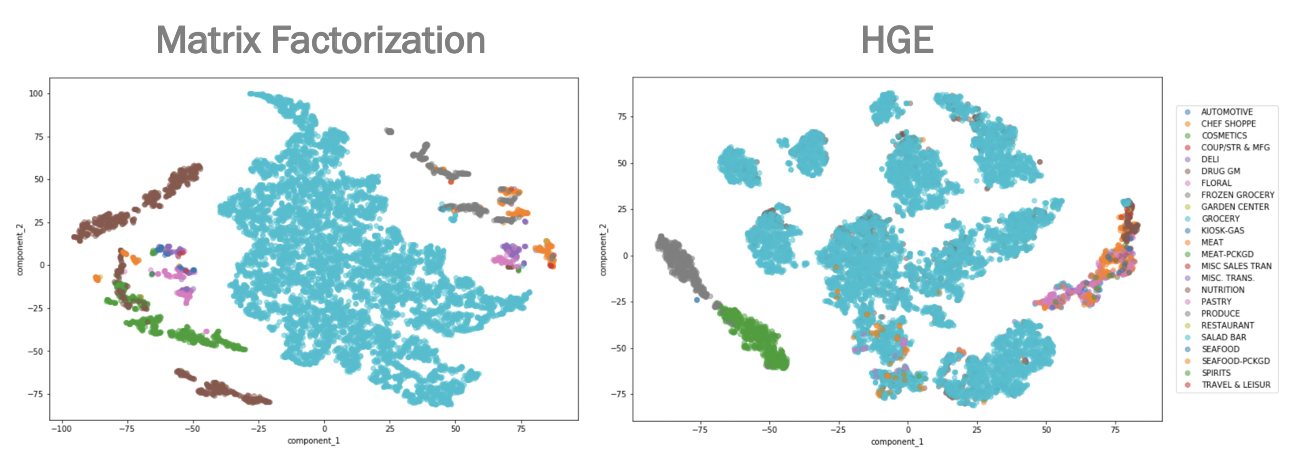}}
\caption{t-SNE item embeddings for the Retail Rocket dataset using two components. Left: Matrix Factorization. Right: HGE}
    \label{fig:5.2:item_embeddings}
\end{center}
\end{figure*}

As we can see from \autoref{fig:5.2:item_embeddings}, we use t-Distributed Stochastic Neighbor Embedding on the item embeddings for the Retail Rocket dataset, \cite{Maaten2008-ce}. 
The item embedding space is more clustered for the HGE model than for MF. 
Moreover, we can mention that inside the most significant category, grocery, there are sub-clusters. These sub-clusters are lower-level categories inside the grocery category. 
Thus, we see that for the Retail Rocket dataset, HGE yields hierarchically clustered item embeddings. 
Also, with HGE, there is a logical structure in the item embedding space even for the highest level categories that canonical MF embeddings do not have. Similar categories are closer to each other, and dissimilarities are distant. 
For example, packaged meat and frozen grocery, categories colored in green and gray, are close to each other in the HGE item embedding space. 
We can elucidate that HGE embeddings reflect the closeness of packaged food with long shelf life.

The more item hierarchy levels available, the better results HGE yields. 
Indeed, for the three most significant datasets with three levels of item hierarchies or more, one to two levels meaning from one to two HGE layers, one per level, respectively, improved the model precision@20 on average by 6\%. 
An increase from two to three levels improved it on average by 8\%. 
Thus, we suggest using HGE for datasets with at least two item hierarchy levels. 
For example, we can use it for datasets in retail and e-commerce.

HGE has more number of parameters than canonical matrix factorization, but only with a constant multiplier. 
So, asymptotically, the number of parameters is the same. 
One could argue that convolution operations may perform slower than the dot product, \cite{gnn_implementation}, so we decided to compare training time for HGE and canonical MF. 
We experimented with the following settings. 
First, we built both models in PyTorch from scratch and trained them on the same hardware. 
Second, we used the same number of iterations and learning rate. 
Third, we compare training time for hidden dimension sizes from 20 to 200 with a step of 20. 
Finally, for these five datasets, HGE is only 14\% slower than the canonical matrix factorization. 
This result is not dependent on the hidden dimension size.

\section {Ablation study} \label{sec:ablation}
\begin{figure}[ht]
    \centering
    \includegraphics[width=0.6\columnwidth]{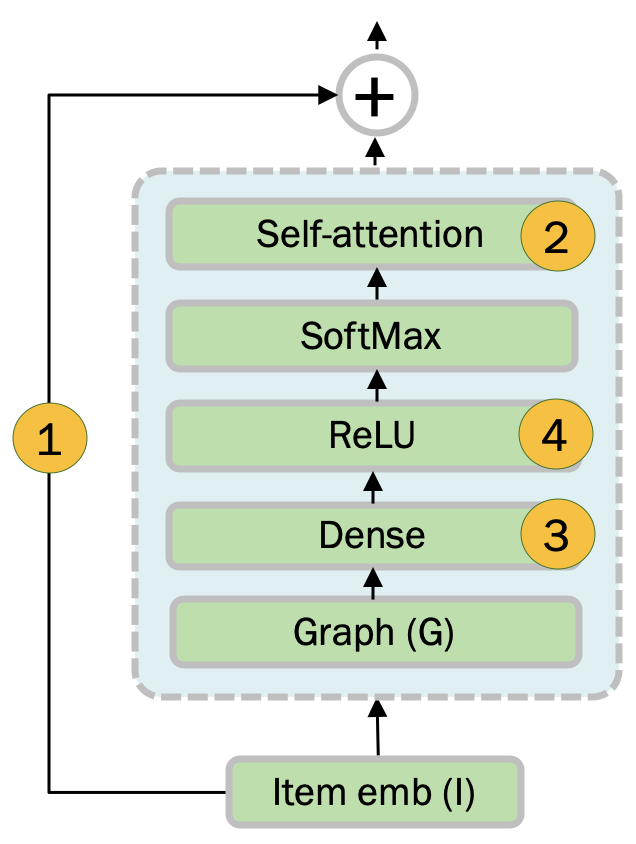}
    \caption{Scheme for the Ablation Study}
    \label{fig:5.3:ablation_study}
\end{figure}

In \autoref{fig:5.3:ablation_study}, we can see the resulting HGE layer structure and the proposed ablation study. 
After conducting it, we list in \autoref{tab:ablation} the averages across our baselines in terms of change in precision@20.

\begin{table*}[ht]
\centering
\caption{\textsc{Ablation Study}\label{tab:ablation}}
\resizebox{\textwidth}{!}{%
\begin{tabular}{llc}
Step & Ablation & Change in precision@20 \\ \hline
(1) &  Removing skip-connection & \textbf{-17.2\%}   \\ \hline
(2) & Adding self-attention  &  \textbf{+0.2\%}   \\ \hline
(3) & Replacing matrix $W$ with its factorization $W_1*W_2^T$  &  \textbf{+1.1\%}   \\ \hline
(4) & Removing ReLU activation  &  \textbf{-6.7\%}   \\ \hline
(4) & Replacing ReLU with Leaky ReLU  &  \textbf{-1.6\%}   \\ \hline
(5) & Replacing skip-connection sum operation with 1x1 convolution over several graph hierarchy levels &  \textbf{-0.4\%}   \\ \hline
(5) & Replacing skip-connection sum operation with attention over several graph hierarchy levels  &  \textbf{+0.7\%}   \\ \hline
\end{tabular}
}
\end{table*}

In point (1), we see that removing the skip-connection results in the most significant drop in model performance. 
It seems logical as most of the graph-based models assume that final item embedding is some function from the item embedding $I_{t-1}$ in the previous state and the embedding of its neighbors $I_{t-1}^{\textrm{neighbors}}$ define as

\begin{equation*}
I_t = f(I_{t-1}; I_{t-1}^{\textrm{neighbors}}).
\end{equation*} 

In our case, the final layer architecture suggests that the aggregation function is a sum function. 
Indeed, assuming that the item embedding consists only of its neighbors' embeddings and is not dependent on its embedding at the previous state, it results in losing valuable information about the item itself and leads to poor performance.

For point (2), several articles claim that adding attention inside the graph convolutional layer may improve its performance, \cite{kgat, dual_graph_attention, gnn_survey}. 
In our study, adding self-attention improves the performance of the model marginally. 
However, attention operation is complex. 
It has a large number of learnable parameters and mathematical operations. 
We tried to create a very lightweight and fast model. 
Hence, we decided to drop this mechanism from the final architecture.

In point (3), to improve the model's speed and the number of parameters, we replaced the matrix $W$ with its factorization $W_1*W_2^T$. 
Let $I$ be the number of items in the dataset, $K$ be the total number of categories at a particular hierarchy level, and $h$ be the hidden layer size. 
Replacing $W$ with its factorization $W_1*W_2^T$ did not decrease model performance but reduced the number of parameters from $I * K$ to $(I+K)*h$. 
The worst-case scenario for $K$ is $K=log_2(I)$. 
It is a case when the item hierarchy graph is a binary tree. 
So, by replacing the matrix with its factorization, we asymptotically reduced the number of parameters from $O(I*log_2(I))$ to $O(I)$, which seems to be a good result.

Point (4) is devoted to a quantitative study on the activation function. 
Removing the ReLU activation function leads to a decrease in performance. 
We assume that ReLU serves as an additional regularization, pushing all negative values from the dense layer to zero. 
Thus, with ReLU, it is possible to zero the impact of some items on the final category embedding. 
This property may reflect many real-world cases. 
Let us consider sales in offline retail, specifically the sweet sparkling water category and the product Coca-Cola in it. 
Coca-Cola is a particular item. 
It is a complementary product to others, such as alcohol or chips. 
So, it is reasonable that the Coca-Cola embedding is close to embeddings of alcohol items. 
However, customers rarely purchase other sweet sparkling water together with alcohol. 
Thus, it seems logical to exclude Coca-Cola from calculating the sparkling water category embedding in the HGE layer. 
We can do it with the help of ReLU. 
Another experiment conducted on point (4) is to replace ReLU with Leaky ReLU. 
We can describe the latter activation function as

\begin{equation*}
\textrm{Leaky ReLU} = \textrm{max} (\alpha x, x),
\end{equation*} 

The coefficients $\alpha > 0$ and $\alpha < 1$ correspond to some small value. 
They let the gradient be non-zero for a negative activation function input, \cite{handbook}. 
Replacing ReLU with Leaky ReLU decreases the model performance in precision@20 and leads to a less sparse activation function output as none of the output is zero.

The last two points (5, 6) correspond to item embeddings and HGE embeddings from several pooling layers. 
Indeed, our final variant does not have pooling. 
We use skip-connection after each HGE layer. 
Adding pooling with $1 \times 1$ convolution or attention did not improve the result. 
Our resnet-like skip-connections proved to be powerful enough to handle several item hierarchy graph layers. 

The ablation study was critical to speed-up the model and to increase its performance. 
From the study, we can conclude the following.

\begin{itemize}
    \item We can safely remove complex parts as self-attention or $1 \times 1$ convolution over several HGE layers without a drop in performance and get additional model speed-up;
    \item The skip-connection is an essential and robust mechanism for remembering item features;
    \item We can get an asymptotical improvement in the number of parameters from $O(I*log_2(I))$ to $O(I)$ by replacing matrix $W$ with its factorization $W_1*W_2^T$;
    \item The ReLU activation serves as an additional regularization. It gives the possibility to zero the impact of some items on the final category embedding.
\end{itemize}

\section{Discussion} \label{sec:discussion}
\begin{figure}[ht]
    \centering
    \includegraphics[width=\columnwidth]{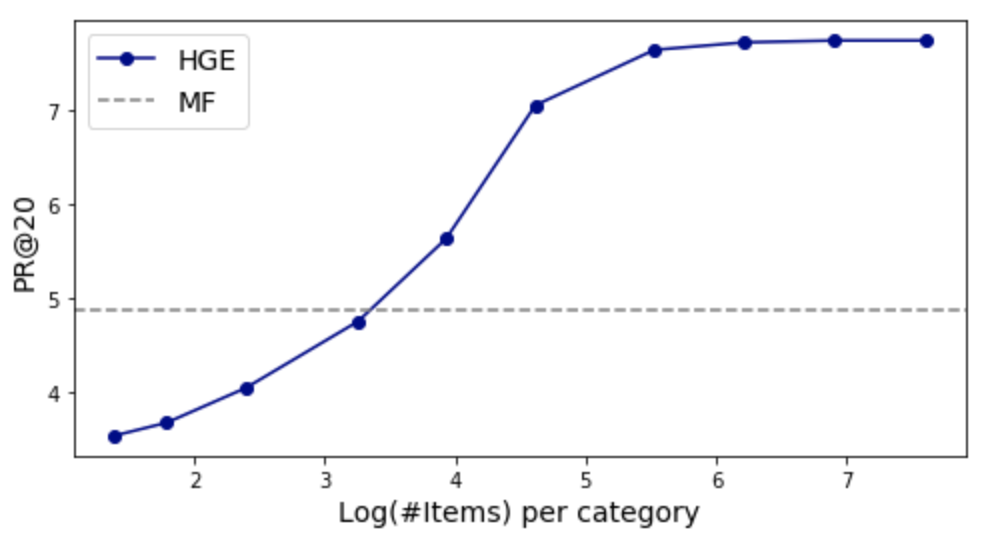}
    \caption{Dependency between precision@20 and log number of items per category}
    \label{fig:5.4:items_pr}
\end{figure}

The main drawback of the proposed model directly refers to fitting the HGE layer. 
Indeed, it requires several items in each category per batch to update the weights of the HGE layer correctly. 
Consequently, if a few items per category are present, the performance of HGE will be moderate, as seen in \autoref{fig:5.4:items_pr}.

Indeed, the HGE yields stable and excellent performance for categories with more than 150 items. 
HGE outperforms the canonical MF for categories with more than 50 items. 
One possibility to obtain broad categories is to drop those with fewer than 150 items. 
Another possibility is to combine such categories into an "Other" category. 
We suggest using HGE only for medium and large-sized datasets as they have enough items per category property due to their size.

\begin{figure}[ht]
    \centering
    \includegraphics[width=\columnwidth]{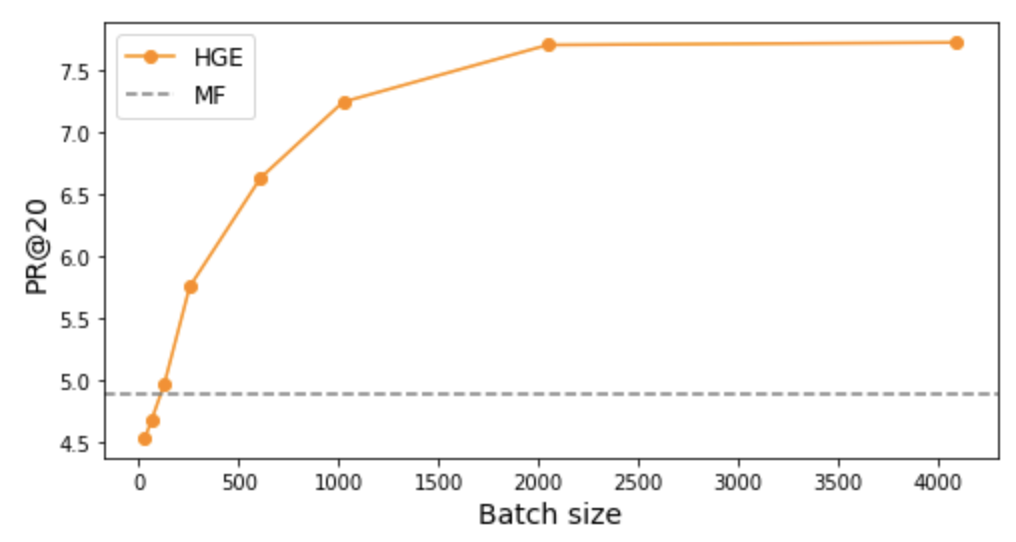}
    \caption{Dependency between precision@20 and batch size}
    \label{fig:5.5:batchsize_pr}
\end{figure}

Nevertheless, even with broad categories, we can still have problems in learning the parameters of HGE. 
Inside the batch of each category, we can have few items. We propose two improvements to it. 
We present the results of the experiment in \autoref{fig:5.5:batchsize_pr}. The first one is a straightforward stratified batch sampling. 
Each batch should contain the same number of items per category or number of items proportional to each category's log size. 
We use log normalization to reduce the impact of significant categories and increase minor categories' impact on the sampling technique. 
The second improvement is to increase the batch size.

The sampling technique works well even with small batch sizes of 128 to 256 compared to the canonical MF. 
However, this technique reaches its plateau of excellent performance, starting from batch sizes of 2056. 
So, to solve the problem of few items per category, we suggest using large batch sizes, which leads to an increase in computational time.

We can, therefore, summarize our recommendations as

\begin{itemize}
    \item Use HGE for medium and large size datasets, see \autoref{sec:ablation});
    \item Use HGE when multiple item hierarchy levels are present, see \autoref{sec:ablation};
    \item Combine categories with less than 150 items into an "Other" category, see \autoref{fig:5.4:items_pr};
    \item Use stratified sampling for batch generation, see \autoref{sec:discussion};
    \item Use a big enough batch size, see \autoref{fig:5.5:batchsize_pr}.
\end{itemize}

\section{Summary \& Future Work} \label{summary}
The study proves that the model with an item hierarchy graph as additional input and a particular hierarchical graph embedding layer can perform exceptionally well in the cold start setting. 
After quantitative experiments on five datasets, we conclude that HGE proved its competitiveness. 
HGE is the best model for medium and large datasets with several item hierarchy levels in all the proposed metrics when compared to baselines, HGE is still robust, and it is not worse than baselines.

In order to make HGE fast and scalable, we introduced matrix factorization inside the HGE layer. 
Theoretically, it decreases the number of hierarchical graph embedding layer parameters from $O(\# items * log_2(\# items))$ to $O(\# items)$. Such asymptotical property makes HGE scalable even for large datasets. 
Moreover, as HGE is an improvement of matrix factorization, alternating least squares optimization can be applied to it, increasing performance even further and making HGE ready for a production environment. 
After several experiments, HGE turned out to be only 14\% slower than the canonical matrix factorization for the five datasets discussed.

Moreover, we addressed the issue of removing some item embeddings from the final graph embedding. 
For this purpose, we adapted the ReLU activation inside the HGE layer. 
This property is crucial for real-world data where multiple undesired outliers in item embeddings may exist. 
Our model is capable of filtering them.

One of the characteristics of HGE is that it requires sufficient items per category and batch. To handle cases where this is not possible, we recommend the following. 
First, HGE shows improvement in performance only for medium, 2M+ interactions, large datasets, and 10M+ interactions. 
Second, HGE requires more than 150 items per category. 
To overcome this issue, we can combine small categories into an "Other" category. 
Third, to fit HGE, we require a big batch size. 
Fourth, HGE training requires stratified sampling by category for the batch creation.

There are multiple options to improve the proposed algorithm further. 
We can consider including unprocessed external features and item hierarchy graph generated from features inside the HGE layer. 
Also, we can address the cold start problem not only for items but also for users. 
Finally, we can increase the model's speed by applying the alternating least squares mechanism. 

\bibliographystyle{IEEEtran}
\bibliography{bib}

\end{document}